%% file: tro_cores.tex
\documentclass[journal]{IEEEtran}
\pdfoutput=1

\usepackage{etex}

\ifCLASSINFOpdf
\else
\fi
\usepackage{times}
\usepackage{mathptmx}

\usepackage[numbers,sort&compress]{natbib}
\usepackage{multicol}
\usepackage[bookmarks=true]{hyperref}

\usepackage{color}
\usepackage{amsmath,theorem,stmaryrd}
\usepackage{amssymb,gensymb,ifsym}

\usepackage{graphics,subfigure} 

\usepackage{float}
\floatstyle{ruled}
\newfloat{spec}{thb}{lop}
\floatname{spec}{Specification}
\usepackage{caption}

\usepackage{algorithm}
\usepackage{algorithmicx}
\usepackage[noend]{algpseudocode}
\usepackage{listings}

\theoremstyle{definition}

\newtheorem{goal}{Problem}

\usepackage{booktabs}
\usepackage{tabulary}

\usepackage{tikz}
\usepackage{tikz-qtree}

\newtheorem{theorem}{Theorem}[section]

\theoremstyle{definition}
\newtheorem{definition}{Definition}

\DeclareMathOperator{\F}{\rotatebox[origin=c]{45}{$\Box$}}
\DeclareMathOperator{\G}{\Box}
\DeclareMathOperator{\X}{\bigcirc}

\DeclareMathAlphabet{\mathcal}{OMS}{cmsy}{m}{n}

\newcommand{\comment}[1]{}
\begin{document}
%
\title{Unsynthesizable Cores -- Minimal Explanations for Unsynthesizable High-Level Robot Behaviors}
%
%
%

\author{Vasumathi Raman$^{1}$ and Hadas Kress-Gazit$^{2}$
\thanks{*V. Raman is supported by STARnet, a Semiconductor Research Corporation program, sponsored by MARCO and DARPA. H. Kress-Gazit is supported by NSF CAREER CNS-0953365 and DARPA N66001-12-1-4250.}
\thanks{V. Raman is with the Department of Computing and Mathematical Sciences, California Institute of Technology, Pasadena CA 91125 (e-mail: vasu@caltech.edu).}%
\thanks{$^{2}$H. Kress-Gazit is with the Sibley School of Mechanical and Aerospace Engineering, 
		Cornell University, Ithaca, NY 14853, USA
        {\tt\small hadaskg at cornell.edu}}%
\thanks{Manuscript received August 21, 2014.}}

\maketitle

\begin{abstract}
\input{abstract}

\end{abstract}


%
\IEEEpeerreviewmaketitle

\section{Introduction}
\label{intro}
\input{intro}

\section{Preliminaries}
\label{prelims}
\input{prelims}

\section{Problem Statement}
\label{problem}
\input{problem}

\section{Related Work}
\label{related}
\input{related}


\section{Unsatisfiable Cores via SAT}
\label{cores_unsat}
\input{cores_unsat}

\section{Unrealizable Cores via SAT}
\label{cores_unreal}
\input{cores_unreal}

\section{Unsynthesizable Cores via Iterated Synthesis}
\label{cores_iterate}
\input{cores_iterate}

\section{Examples}
\label{examples}
\input{examples}

\section{Conclusions}
\label{conclusions}
\input{conclusions}


%

%


\ifCLASSOPTIONcaptionsoff
  \newpage
\fi



%
\bibliographystyle{IEEEtran}
\bibliography{biblio}
%





\end{document}

%% file: abstract.tex
With the increasing ubiquity of multi-capable, general-purpose robots arises the need for enabling non-expert users to command these robots to perform complex high-level tasks. To this end, high-level robot control has seen the application of formal methods to automatically synthesize correct-by-construction controllers from user-defined specifications; synthesis fails if and only if there exists no controller that achieves the specified behavior. Recent work has also addressed the challenge of providing easy-to-understand feedback to users when a specification fails to yield a corresponding controller. Existing techniques provide feedback on portions of the specification that cause the failure, but do so at a coarse granularity. This work presents techniques for refining this feedback, extracting minimal explanations of unsynthesizability.

%% file: intro.tex
As robots become increasingly general-purpose and ubiquitous, there is a growing need for them to be easily controlled by a wide variety of users. The near future will likely see robots in homes and offices, performing everyday tasks such as fetching coffee and tidying rooms. The challenge of programming robots to perform these tasks has until recently been the domain of experts, requiring hard-coded implementations with the ad-hoc use of low-level techniques such as path-planning during execution. 

Recent advances in the use of formal methods for robot control have enabled non-expert users to command robots to perform high-level tasks using a specification language instead of programming the robot controller (e.g., \cite{Kloetzer2008a,Karaman09,Bhatia2010,LaValle11,KGF,Nok10}). There are several approaches in which correct-by-construction controllers are automatically synthesized from a description of the desired behavior and assumptions on the environment in which the robot operates \cite{KGF,Nok10}. If a controller implementing the specification exists, one is returned. However, for specifications that have no implementation (i.e. \emph{unsynthesizable} specifications), the process of pin-pointing the cause of the problem can be a frustrating and time-consuming process. This has motivated recent algorithms for explaining unsynthesizability of specifications \cite{ICRA12,TRO12}, revising specifications \cite{Fainekos11}, and adding environment assumptions that would make the specification synthesizable \cite{Li11}. 

There are two ways in which a specification can be unsynthesizable -- it is either \emph{unsatisfiable}, in which case the specified robot behavior cannot be achieved in \emph{any} environment, or it is \emph{unrealizable}, in which case there exists an admissible environment (satisfying the specified assumptions) that prevents the robot from achieving its specified behavior. 
Feedback about the cause of unsynthesizability can be provided to the user in the form of a modified specification \cite{Fainekos11,KimFainekos12a,KimFainekos12b}, a highlighted fragment of the original specification \cite{CAV11}, or by allowing the user to interact with a simulated adversarial environment that prevents the robot from achieving the specified behavior \cite{TRO12}. 

Previous approaches left open the challenge of refining feedback to the finest possible granularity, providing the user with a \emph{minimal} cause of unsynthesizability. The main contribution of this paper is to to identify \emph{unsynthesizable cores} -- minimal subsets of the desired robot behavior that cause it to be unsatisfiable or unrealizable. The analysis makes use of off-the-shelf Boolean satisfiability (SAT) solvers and existing synthesis tools to find this minimal cause of unsynthesizability, and therefore lends itself to generalization to any formal specification language for which the relevant tools exist. This paper subsumes the results presented in \cite{RSS13,IROS13}, and extends the core-finding capabilities to previously unaddressed cases, as described in Section \ref{cores_iterate}. In particular, this is the first paper to present a sound and complete algorithm for finding unsynthesizable cores (as defined in Section \ref{problem}) for specifications in the GR(1) fragment of Linear Temporal Logic (LTL), and discuss special considerations necessitated by specifications in the robotics problem domain.

The paper is structured as follows. Section \ref{prelims} reviews terms and preliminaries. Section \ref{problem} describes types of unsynthesizability, and presents a formal definition of the problem of identifying unsynthesizable cores. Section \ref{related} describes related work on analyzing unsynthesizable specifications. Sections \ref{cores_unsat} and \ref{cores_unreal} present techniques for using Boolean satisfiability to identify unsatisfiable and unrealizable cores respectively. Section \ref{cores_iterate} describes an alternative method for identifying cores, based on iterated realizability checks. Section \ref{examples} demonstrates the effectiveness of the more fine-grained feedback on example specifications. The paper concludes with a description of future work in Section \ref{conclusions}.

%% file: prelims.tex
The high-level tasks considered in this work involve a robot operating in a known workspace. The robot reacts to events in the environment, which are captured by its sensors, in a manner compliant with the task specification, by choosing from a set of actions including moving between adjacent locations. The tasks may include infinitely repeated behaviors such as patrolling a set of locations. Examples of such high-level tasks include search and rescue missions and the DARPA Urban Challenge.

\subsection{Controller Synthesis}
High-level control for robotics is inherently a hybrid domain, consisting of both discrete and continuous components. Automated correct-by-construction controller synthesis for this domain using formal methods requires a discrete abstraction and a description of the task in a formal specification language. The discrete abstraction of the high-level robot task consists of a set of propositions ${\cal X}$ whose truth value is controlled by the environment and read by the robot's sensors, and a set of action and location propositions $\cal Y$ controlled by the robot; the set of all propositions $AP = {\cal X} \cup {\cal Y}$. The value of each $\pi \in AP$ is the abstracted binary state of a low-level black box component. More details on the discrete abstraction used in this work can be found in \cite{KGF}. 

The formal language used for high-level specifications in this work is Linear Temporal Logic (LTL) \cite{LTL}, a modal logic that includes temporal operators, allowing formulas to specify the truth values of atomic propositions  over time. LTL is appropriate for specifying robotic behaviors because it provides the ability to describe changes in the truth values of propositions over time. To allow users who may be unfamiliar with LTL to define specifications, some tools like LTLMoP \cite{LTLMoP} include a parser that automatically translates English sentences belonging to a defined grammar \cite{grammar} into LTL formulas, as well as some natural language capabilities, as described in \cite{RSS13}. 

\subsubsection{Linear Temporal Logic (LTL)}
LTL formulas are constructed from atomic propositions $\pi \in AP$ according to the following recursive grammar:
$$\varphi ::= \pi \mid \neg \varphi \mid\varphi \vee \varphi \mid \X \varphi \mid \varphi~{\mathcal U}~\varphi,$$
where $\neg$ is negation, $\vee$ is disjunction, $\X$ is ``next'', and ${\mathcal U}$ is a strong ``until". 
Conjunction ($\wedge$), implication ($\Rightarrow$), equivalence ($\Leftrightarrow$), ``eventually" ($\F$) and 
``always" ($\G$) are derived from these operators. The truth of an LTL formula is evaluated over sequences of truth assignments to the propositions in $AP$. In this paper, truth assignments are represented as subsets  $P \subseteq AP$, with $\pi \in P$ being set to true and $\pi \in AP\backslash P$ being false. Informally, the formula $\X \varphi$ expresses that $\varphi$ is true in the next position in the sequence. Similarly, $\G \varphi$ expresses that $\varphi$ is true at every position, and $\F\varphi$ expresses that $\varphi$ is true at some position in the sequence. Therefore, the formula $\G\F\varphi$ is satisfied if $\varphi$ is true infinitely often. For a formal definition of the semantics of LTL, see \cite{MCBk}. 


The task specifications in this paper are expressed as LTL formulas of the form $\varphi = \varphi_e \Rightarrow \varphi_s$, where $\varphi_e$ encodes assumptions about the environment's behavior and $\varphi_s$ represents the desired robot behavior. $\varphi_e$ and $\varphi_s$ each have the structure $\varphi_p = \varphi_p^i \land \varphi_p^t \land \varphi_p^g$, where $\varphi_p^i, \varphi_p^t$ and $\varphi_p^g$ for $p \in \{e,s\}$ represent the initial conditions, transition relation and goals for the environment ($e$) and the robot ($s$) respectively. This fragment of LTL is called Generalized Reactivity (1) or GR(1) \cite{Piterman06}. 

The subformulas $\varphi^t_e$ and $\varphi^t_s$ above are referred to as \textit{safety} formulas, and encode assumptions on the environment and restrictions on the system transitions respectively. Each consists of a conjunction of formulas of the form $\G A_i$, where each $A_i$ is a Boolean formula over $AP$ (formulas over $\X AP = \{\X \pi \mid \pi \in AP\}$ are also allowed in $\varphi^t_s$). On the other hand, $\varphi_g^e$ and $\varphi_g^s$ are referred to as \textit{liveness} formulas, and consist of conjunctions of clauses of the form $\G \F B_j$. Each $B_j$ is a Boolean formula over $AP$, and represents an event that should occur infinitely often when the robot controller is executed. The initial conditions $\varphi_e^i$ and $\varphi_s^i$ are non-temporal Boolean formulae over $\cal X$ and $\cal Y$ respectively.

An LTL formula $\varphi$ is \emph{realizable} if, for every time step, given a truth assignment to the environment propositions for the next time step, there is an assignment of truth values to the robot propositions such that the resulting infinite sequence of truth assignments satisfies $\varphi$. The synthesis problem is to find an automaton that encodes these assignments, i.e. whose executions satisfy $\varphi$. For a synthesizable specification $\varphi$, synthesis produces an implementing automaton, enabling the construction of a hybrid controller that produces the desirable high-level, autonomous robot behavior. The reader is referred to \cite{Piterman06} and \cite{KGF} for details of the synthesis procedure, and to \cite{KGF,LTLMoP} for a description of how the extracted discrete automaton is transformed into low-level robot control. 

\subsection{Environment Counterstrategy}


In the case of unsynthesizable specifications, the counterstrategy synthesis algorithm introduced in \cite{Konig09} gives an automatic method of constructing a strategy for the environment, which provides sequences of environment actions that prevent the robot from achieving the specified behavior. The counterstrategy takes the form of a finite state machine:

\begin{definition}
An \emph{environment counterstrategy} for LTL formula $\varphi$ is a tuple $A^e_\varphi = (Q, Q_0, {\cal X}, {\cal Y}, \delta_e, \delta_s, \gamma_{\cal X}, \gamma_{\cal Y}, \gamma_{goals})$ where 
\begin{itemize}
\item $Q$ is a set of states. 
\item $Q_0 \subseteq Q$ is a set of initial states.
\item ${\cal X}$ is a set of inputs (sensor propositions).
\item ${\cal Y}$ is a set of outputs (location and action propositions).
\item $\delta_e:Q \rightarrow 2^{\cal X}$ is a deterministic input function, which provides the input propositions that are true in the next time step given the current state $q$, and satisfies $\varphi_e^t$.
\item $\delta_s:Q \times 2^{\cal X} \rightarrow 2^{Q}$ is the (nondeterministic) transition relation. If $\delta_s(q,x) = \emptyset$ for some $x \in 2^{\mathcal X}, q \in Q$, then there is no next-step assignment to the set of outputs that satisfies the robot's transition relation $\varphi_s^t$, given the next set of environment inputs $x$ and the current state $q$.
\item $\gamma_{\cal X}:Q \rightarrow 2^{\cal X}$ is a transition labeling, which associates with each state the set of environment propositions that are true over incoming transitions for that state (note that this set is the same for all transitions into a given state). Moreover, if $q' \in \delta_s(q,x)$ then $\gamma_{\cal X}(q') = x$. 
\item $\gamma_{\cal Y}:Q \rightarrow 2^{\cal Y}$ is a state labeling, associating with each state the set of robot propositions true in that state.
\item $\gamma_{goals}: Q \rightarrow \mathbb{Z^+}$ labels each state with the index of a robot goal that is prevented by that state. During the counterstrategy extraction, every state in the counterstrategy is marked with some robot goal \cite{Konig09}.
\end{itemize}
\end{definition}

The counterstrategy $A^e_\varphi$ provides truth assignments to the input propositions (in the form of the transition function $\delta_e$) that prevent the robot from fulfilling its specification. 
The inputs provided by $\delta_e$ in each state satisfy $\varphi_e^t$, meaning that for all $q \in Q$, the truth assignment sequence $(\gamma_{\cal X}(q)\cup\gamma_{\cal Y}(q), \gamma_{\cal X}(\delta_e(q)))$ satisfies $A_i$ for each conjunct $A_i$ in $\varphi_e^t$ (note that $A_i$ is a formula over two consecutive time steps). In addition, all infinite executions of $A^e_\varphi$ satisfy $\varphi_e \land \neg \varphi_s$.

%% file: problem.tex
A specification that does not yield an implementing automaton is called \emph{unsynthesizable}. Unsynthesizable specifications are either \emph{unsatisfiable}, in which case the robot cannot succeed no matter what happens in the environment (e.g., if the task requires patrolling a disconnected workspace), or \emph{unrealizable}, in which case there exists at least one environment that can prevent the desired behavior (e.g., if in the above task, the environment can disconnect an otherwise connected workspace, such as by closing a door). More examples illustrating the two cases can be found in \cite{TRO12}. 

In either case, the robot can fail in one of two ways: either it ends up in a state from which it has no moves that satisfy the specified safety requirements $\varphi_s^t$ (this is termed \emph{deadlock}), or the robot is able to change its state infinitely, but one of the goals in $\varphi_s^g$ is unreachable without violating $\varphi_s^t$ (termed \emph{livelock}). In the context of unsatisfiability, an example of deadlock is when the system safety conditions contain a contradiction within themselves. Similarly, unrealizable deadlock occurs when the environment has at least one strategy for forcing the system into a deadlocked state.  Livelock occurs when there is one or more goals that cannot be reached while still following the given safety conditions. 

\begin{figure}
\centering
\includegraphics[width=.49\textwidth]{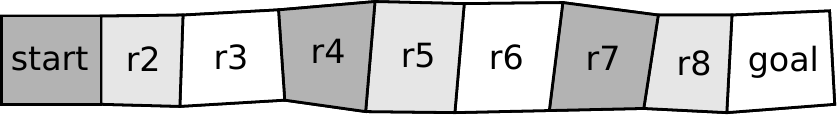}
\caption{Map of robot workspace in Specification \ref{spec:unreal1}}
\label{fig:hallway}
\end{figure}

Consider Specification~\ref{spec:unreal1}, in which the robot is operating in the workspace depicted in Fig. \ref{fig:hallway}. The robot starts at the left hand side of the hallway (\ref{item:a0}), and must visit the goal on the right (\ref{item:a3}). The safety requirements specify that the robot should not pass through region $r5$ if it senses a person (\ref{item:a1}). Additionally, the robot should always activate its camera (\ref{item:a2}).

\begin{spec}
\small
\begin{enumerate}
\item \textit{Robot starts in start with camera} \\($\pi_{start} \land \pi_{camera}$,  part of $\varphi_s^i)$) \label{item:a0}
\item \textit{If you are sensing a person then do not r5} \\($\G (\X \pi_{person} \Rightarrow \neg \X \pi_{r5})$,  part of $\varphi_s^t$) \label{item:a1}
\item \textit{Always activate the camera} ($\G \X \pi_{camera}$, part of $\varphi_s^t$) \label{item:a2}
\item \textit{Visit the goal} ($\G\F \pi_{goal}$, part of $\varphi_s^g$) \label{item:a3}
\end{enumerate}
\caption{Unrealizable specification -- livelock}
\label{spec:unreal1}
\end{spec}

It is clear that the environment can prevent the goal in (\ref{item:a3}) by always activating the ``person'' sensor ($\pi_{person}$), because of the initial condition in (\ref{item:a0}) and the safety requirement in (\ref{item:a1}). Note that Specification~\ref{spec:unreal1} is a case of livelock: the robot can satisfy the safety requirement indefinitely by moving between the first four rooms on the left, but is prevented from ever reaching the goal if it sees a person all the time -- the environment is able to disconnect the topology using the person sensor. 

Previous work produced explanations of unsynthesizability in terms of combinations of the specification components (i.e., initial conditions, safeties and goals) \cite{TRO12}. However, in many cases, the true conflict lies in small subformulas of these components. For example, the safety requirement (\ref{item:a2}) in Specification~\ref{spec:unreal1} is irrelevant to its unsynthesizability, and should be excluded from any explanation of failure. The specification analysis algorithm presented in \cite{TRO12} will narrow down the cause of unsynthesizability to the goal in (\ref{item:a3}), but will also highlight the entirety of $\varphi_s^t$, declaring that the environment can prevent the goal because of some subset of the safeties (without identifying the exact subset).

This motivates the identification of small, minimal, ``core'' explanations of the unsynthesizability. A first step is to define what is meant by an \emph{unsynthesizable core}. This paper draws inspiration from the Boolean satisfiability (SAT) literature to define an unsynthesizable core of a GR(1) LTL formula.  Given an unsatisfiable SAT formula in conjunctive normal form (CNF), an unsatisfiable core is traditionally defined as a subset of CNF clauses that is still unsatisfiable. A \emph{minimal} unsatisfiable core is one such that every proper subset is satisfiable; a given SAT formula can have multiple minimal unsatisfiable cores of varying sizes. This definition should be distinguished from that of a \emph{minimum} unsatisfiable core, which is one containing the smallest number of original clauses that are unsatisfiable in themselves. While there are several practical techniques for computing minimal unsatisfiable cores, and many modern SAT-solvers include this functionality, there are no known practical algorithms for computing minimum cores. This paper will therefore focus on leveraging existing tools for computing minimal unsatisfiable cores, to compute \emph{minimal unsynthesizable cores}.

Let $\varphi_1 \preceq \varphi_2$ ($\varphi_1 \prec \varphi_2$) denote that $\varphi_1$ is a subformula (strict subformula) of $\varphi_2$.

\begin{definition}
Given a specification $\varphi = \varphi_e \Rightarrow \varphi_s$, a \emph{minimal unsynthesizable core} is a subformula $\varphi^*_s \preceq \varphi_s$ such that $\varphi_e \Rightarrow \varphi^*_s$ is unsynthesizable, and for all $\varphi'_s \prec \varphi^*_s$, $\varphi_e \Rightarrow \varphi'_s$ is synthesizable.
\end{definition}
\begin{goal}
Given an unsynthesizable formula $\varphi$, return a minimal unsynthesizable core $\varphi^*_s \preceq \varphi_s$.
\end{goal}

%% file: related.tex
Robotics researchers have recently considered the problem of revising specifications that cannot be satisfied by a given robot system \cite{Fainekos11,KimFainekos12a,KimFainekos12b} . The work in \cite{Fainekos11} addressed the problem of revising unsatisfiable LTL specifications. The author defines a partial order on LTL formulas, and defines the notion of a valid relaxation for an LTL specification, which informally corresponds to the set of formulas ``greater than'' formulas in the specification according to this partial order. Formula relaxation for unreachable states is accomplished by recursively removing all positive occurrences of unreachable propositions. Specifications with logical inconsistencies are revised by augmenting the synchronous product of the robot and environment specifications with previously disallowed transitions as needed to achieve the goal state. In \cite{KimFainekos12a,KimFainekos12b}, the authors present exact and approximate algorithms for finding minimal revisions of specification automata, by removing the minimum number of constraints from the unsatisfiable specification. They too encode the problem as an instance of Boolean satisfiability, and solve it using efficient SAT solvers. However, the work presented in this paper differs in its objective, which is to provide feedback on existing specifications, not rewrite them. Moreover, unlike the above approaches, this work deals with reactive specifications.

Although explaining unachievable behaviors has only recently been studied in the context of robotics, there has been considerable prior work on unsatisfiability and unrealizability of LTL in the formal methods literature, and the problem of identifying small causes of failure has been studied from several perspectives. For unsatisfiable LTL formulas, the authors of \cite{Schuppan09} suggest a number of notions of unsatisfiable cores, tied to the corresponding method of extraction. These include definitions based on the syntactic structure of the formula parse tree, subsets of conjuncts in various conjunctive normal form translations of the formula, resolution proofs from bounded model-checking (BMC), and tableaux constructions. The authors of \cite{Beer12} employ a formal definition of causality to explain counterexamples provided by model-checkers on unsatisfiable LTL formulas; the advantage of this method is the flexibility of defining an appropriate causal model.

The technique of extracting an unsatisfiable core from a BMC resolution proof is one that is well-used in the Boolean satisfiability (SAT) and SAT Modulo Theories (SMT) (e.g., \cite{SATCores,SMTCores}) literature. A similar technique was used in \cite{Shlyakhter03} for debugging declarative specifications. In that work, the abstract syntax tree (AST) of an inconsistent specification was translated to CNF, an unsatisfiable core was extracted from the CNF, and the result was mapped back to the relevant parts of the AST. The approach in \cite{Shlyakhter03} only generalizes to specification languages that are reducible to SAT, a set which does not include LTL; this paper presents a similar approach, using SAT solvers to identify unsatisfiable cores for LTL. 

The authors of \cite{Cimatti07} also attempted to generalize the idea of unsatisfiable cores to the case of temporal logic using SAT-based bounded model checkers.
Temporal atoms of the original LTL specification were associated with activation variables, which were then used to augment the formulas used by a SAT-based bounded model checker. The result, in the case of an unsatisfiable LTL formula, was a subset of the activation variables corresponding to the atoms that cannot be satisfied simultaneously. The approach presented here for unsatisfiability is very similar, in that the SAT formulas used to determine the core are exactly those that would be used for bounded model checking. However, a major difference is that this work does not use activation variables in order to identify conjuncts in the core, but maintains a mapping from the original formula to clauses in the SAT instance.

In the context of unrealizability, the authors of \cite{Cimatti08} propose definitions for \emph{helpful} assumptions and guarantees, and compute minimal explanations of unrealizability (i.e., \emph{unrealizable cores}) by iteratively expelling unhelpful constraints. Their algorithm assumes an external realizability checker, which is treated as a black box, and performs iterated realizability tests. This work will draw on the same iterative realizability approach in Section \ref{cores_iterate}. The authors in \cite{Konighofer10} use model-based diagnosis to remove not only guarantees but also irrelevant output signals from the specification. These output signals are those that can be set arbitrarily without affecting the unrealizability of the specification. Model-based diagnoses provide more information than a single unrealizable core, but requires the computation of many unrealizable cores. In \cite{Konighofer10}, this is accomplished using techniques similar to those in \cite{Cimatti08}, which in turn require many realizability checks. The main advantage of the work presented here is that it reduces the number of computationally expensive realizability checks required for most specifications, as detailed in Sections \ref{cores_unsat} and \ref{cores_unreal}.

To identify and eliminate the source of unrealizability, some works like \cite{Li11,Chatterjee08} provide a minimal set of additional environment assumptions that, if added, would make the specification realizable; this is accomplished in \cite{Chatterjee08} using efficient analysis of turn-based probabilistic games, and in \cite{Li11} by mining the environment counterstrategy. On the other hand, the work presented in this paper takes the environment assumptions as fixed, and the goal is to compute a minimal subset of the robot guarantees that is unrealizable. Seen from another perspective, this work presumes that the assumptions accurately capture the specification designer's understanding of the robot's environment, and provides the source of failure in the specified guarantees.

%% file: cores_unsat.tex
This section describes how \emph{unsatisfiable} components of the robot specification $\varphi_s$ are further analyzed to narrow the cause of unsatisfiability, for both deadlock and livelock, using Boolean satisfiability testing. Extending these techniques to the environment assumptions $\varphi_e$ is straightforward.

The Boolean satisfiability problem or SAT is the problem of determining whether there exists a truth assignment to a set of propositions that satisfies a given Boolean formula. A Boolean formula in Conjunctive Normal Form (CNF) is one that has been rewritten as a conjunction of clauses, each of which is a disjunctions of literals, where a literal is a Boolean proposition or its negation. For a Boolean formula in CNF, an unsatisfiable core is defined as a subset of CNF clauses whose conjunction is still unsatisfiable; a minimal unsatisfiable core is one such that removing any clause results in a satisfiable formula.

\subsection{Unsatisfiable Cores for Deadlock}
Given a depth $d$ and an LTL safety formula $\varphi$ over propositions $\pi \in AP$, the propositional formula $\psi^d$ is constructed over $\bigcup_{0 \le i \le d+1}AP^i$, where  $\pi^i \in AP^i$ represents the value of $\pi \in AP$ at time step $i$, as:
$$\psi^d(\varphi) = \bigwedge_{0 \le i \le d} \varphi[\X \pi/\pi^{i+1}][\pi/\pi^i],$$ where $\varphi[a/b]$ represents $\varphi$ with all occurrences of subformula $a$ replaced with $b$. This formula is called the depth-$d$ \emph{unrolling} of $\varphi$. Consider Specification~\ref{spec:unsat2}. $\psi^0(\varphi_s^t) = \neg \pi_{kitchen}^0 \land \pi_{camera}^1$ and $\psi^1(\varphi_s^t) = \neg \pi_{kitchen}^0 \land \pi_{camera}^1 \land \neg \pi_{kitchen}^1 \land \pi_{camera}^2$, where $\pi_{kitchen}^i$ is a propositional variable representing the value of $\pi_{kitchen}$ at time step $i$. Given the depth-$d$ unrolling $\psi^d(\varphi_s^t)$ of the robot safety formula, define $\psi^d_{fromInit} = \varphi_s^i[\pi/\pi^0] \land \psi^d(\varphi_s^t)$.

In the case of deadlock, which can be identified as in \cite{ICRA12,TRO12}, a series of Boolean formulas $\{\psi^d_{fromInit}\}$ is produced by incrementally unrolling the robot safety formula $\varphi_s^t$, and the satisfiability of $\psi^d_{fromInit}$ is checked at each depth. To perform this check, the formula $\psi^d_{fromInit}$ is first converted into CNF, so that it can be provided as input to an off-the-shelf SAT-solver; this work uses PicoSAT \cite{PicoSAT}. Converting a Boolean formula to CNF form can, in the worst case, cause an exponential increase in the size of the formula. However, since $\psi^d_{fromInit}$ consists of conjunctions of simple Boolean formulas, the resulting CNFs are small in practice. If $\psi^d_{fromInit}$ is found unsatisfiable, there is no valid sequence of actions that follow the robot safety condition for $d$ time steps starting from the initial condition. In this case, the SAT solver returns a minimal unsatisfiable subformula, in the form of a subset of the CNF clauses. 

When translating the Boolean formula $\psi^d_{fromInit}$ to CNF, a mapping is maintained between the portions of the the original specification, and the clauses they generate. This enables the CNF minimal unsatisfiable core to be traced back to the corresponding safety conjuncts and initial conditions in the specification. 

\begin{spec}
\small
\begin{enumerate}
\item Start in the kitchen ($\varphi_s^i$):\\
\label{item:b1}
$\pi_{kitchen}$

\item Avoid the kitchen ($\varphi_s^i$, $\varphi_s^t$):\\
\label{item:b2}
$\neg \pi_{kitchen} \land \G \neg \pi_{kitchen}$

\item Always activate your camera ($\varphi_s^t$):\\
\label{item:b3}
$\G \X \pi_{camera}$
\end{enumerate}
\caption{Core-finding example -- unsatisfiable deadlock}
\label{spec:unsat2}
\end{spec}

\begin{figure}
\centering
\includegraphics[width=.5\linewidth]{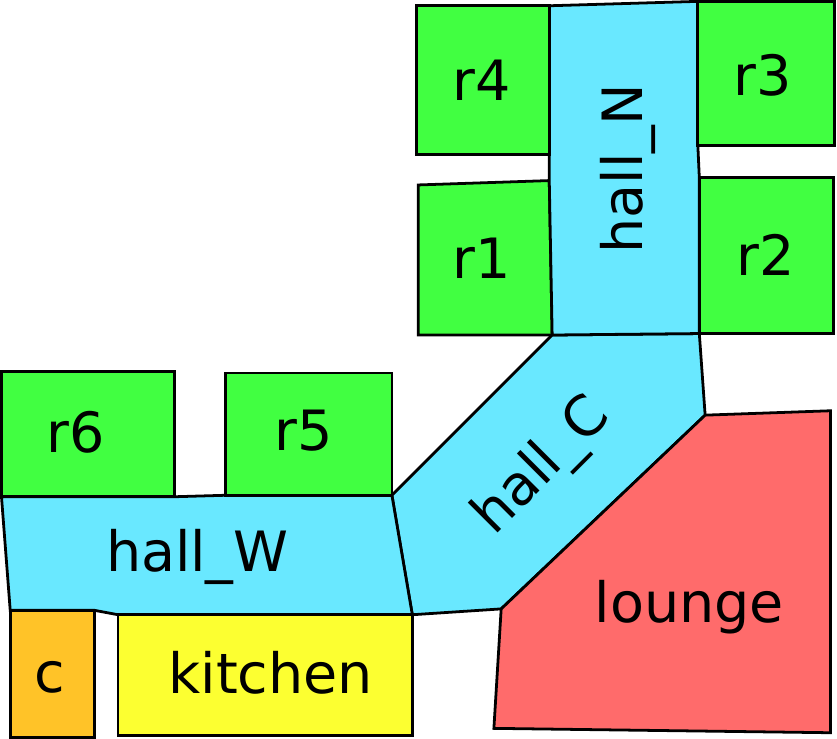}
\caption{Map of hospital workspace (``c'' is the closet)}
\label{fig:hospital_map}
\end{figure}

Specification~\ref{spec:unsat2} is a deadlocked specification, referring to a robot operating in the workspace depicted in Fig. \ref{fig:hospital_map}. The described method begins at the initial state described by $\varphi_s^i$ (lines~\ref{item:b1} and \ref{item:b2}), and unrolls it to $\psi^0_{fromInit} = \pi_{kitchen}^0 \land \neg \pi_{kitchen}^0 \land \neg \pi_{kitchen}^0 \land \pi_{camera}^1$ above. Note that $\psi^0_{fromInit}$ is already unsatisfiable, and the core is given by the subformula $\pi_{kitchen}^0 \land \neg \pi_{kitchen}^0$, which in turn maps back to lines~\ref{item:b1} and \ref{item:b2}. This is because the two statements combined require the robot to both start in the kitchen and not start in the kitchen. Section \ref{examples} contains another example demonstrating unsatisfiable core-finding for deadlock.
  
\subsection{Unsatisfiable Cores for Livelock}
In the case of livelock, a similar unrolling procedure can be applied to determine the core set of clauses that prevent a goal from being fulfilled. A propositional formula is generated by unrolling the robot safety from the initial state for a pre-determined number of time steps, with an additional clause $\varphi^d_{goal}$ representing the unsatisfied liveness condition being required to hold at the final time step for that depth. Consider the livelocked Specification~\ref{spec:unsat3}.

\begin{spec}
\small
\begin{enumerate}
\item Start in the kitchen ($\varphi_s^i$):\\
\label{item:c1}
$\pi_{kitchen}$
\item Avoid hall\_w ($\varphi_s^i$, $\varphi_s^t$):\\
\label{item:c2}
$\neg \pi_{hall}\_w \land \G \neg \pi_{hall}\_w$

\item Always activate your camera ($\varphi_s^t$):\\
\label{item:c3}
$\G \X \pi_{camera}$

\item Patrol r3 ($\varphi_s^g$):\\
\label{item:c3}
$\G \F \pi_{r3}$
\end{enumerate}
\caption{Core-finding example -- unsatisfiable livelock}
\label{spec:unsat3}
\end{spec}

Unrolling the robot safety to depth $d$, with the added clause $\varphi^d_{goal} = \pi_{r3}^d$ for liveness at depth $d$, results in:\begin{eqnarray*}
\psi^d_{fromInit} \land \varphi^d_{goal} &=& \pi_{kitchen}^0 \land \bigwedge_{0 \le i \le d} \neg \pi_{hall\_w}^i  \\
&&\land \bigwedge_{1 \le i \le d+1} \pi_{camera}^i  \land \bigwedge_{0 \le i \le d} \varphi_{topo}^i \land \pi_{r3}^d \
\end{eqnarray*}
\noindent
where $\varphi_{topo}^i$ represents the topology constraints on the robot unrolled at time $i$. $\psi^d_{fromInit}$ is unsatisfiable for any $d \ge 0$. 

\subsection{Unroll Depth}
In the case of deadlock, the propositional formula $\psi^d_{fromInit}$ can be built for increasingly larger depths until it is found to be unsatisfiable for some $d$; by the definition of deadlock, there will always exist such a $d$. This gives us a sound and complete method for determining the depth to which the safety formula must be unrolled in order to identify an unsatisfiable core for deadlock. For livelock, on the other hand, determining the shortest depth that will produce a meaningful core is a much bigger challenge. Consider the above example. For unroll depths less than or equal to 3, the unsatisfiable core returned will include just the environment topology, since the robot cannot reach r3 from the kitchen in 3 steps or fewer, even if it is allowed into $hall\_w$; however, this is not a meaningful core. 

For $d \ge 3$ the core is given by the subformula: $$\pi_{kitchen}^0 \land \bigwedge_{0 \le i \le d} \neg \pi_{hall\_w}^i \land \pi_{r3}^d \land \bigwedge_{0 \le i \le d} \varphi_{topo}^i,$$ which maps back to specification sentences (\ref{item:c1}), (\ref{item:c2}) and (\ref{item:c3}) in Specification \ref{spec:unsat3}. This is because the robot cannot reach $r3$ without passing through $hall\_w$. Section \ref{examples} contains another example demonstrating unsatisfiable core-finding for livelock.

The depth required to produce a meaningful core for unsatisfiability is bounded above by the number of distinct states that the robot can be in, i.e. the number of possible truth assignments to all the input and output propositions. However, efficiently determining the shortest depth that will produce a meaningful core remains a future research challenge, and for the purpose of this work, a fixed depth of 15 time steps was used for the examples presented, unless otherwise indicated.

 Algorithm \ref{alg:unsat_bmc} summarizes the core-finding procedure described in this section. The module {\tt SAT\_SOLVER} takes as input a Boolean formula in CNF form and returns a minimal unsatisfiable core (MUS). {\tt MAP\_BACK} maps the returned CNF clauses to portions of the original specification; in LTLMoP, this mapping returns sentences in structured English or natural language.

\begin{algorithm}
\caption{Unsatisfiable Cores via SAT solving} \label{alg:unsat_bmc}
\begin{algorithmic}[1]
\small
\makeatletter\setcounter{ALG@line}{0}\makeatother
\Function{\tt UNSAT\_BMC}{$\varphi, max\_depth$, reason}
	\State  MUS $\gets \emptyset$
	\If {reason==deadlock}
	\For {$d := 1$ to {max\_depth}}
	\State  MUS $\gets$ {\tt SAT\_SOLVER}($\psi^d_{fromInit}$)
	\If {MUS $\ne \emptyset$}
	\Return {\tt MAP\_BACK}(MUS)
	\EndIf
         \EndFor
         \Else
          \State  MUS $\gets$ {\tt SAT\_SOLVER}($\psi^{max\_depth}_{fromInit} \land \varphi^{max\_depth}_{goal}$)
	 \EndIf
    \Return {\tt MAP\_BACK}(MUS)
 \EndFunction
\end{algorithmic}  
\end{algorithm}


\subsection{Interactive Exploration of Unrealizable Tasks}
\label{mopsy}
If the specification is unrealizable rather than unsatisfiable, the above techniques do not apply directly to identify a core. This is because if  the specification is satisfiable but unrealizable, there exist sequences of truth assignments to the input variables that allow the system requirements to be met. Therefore, in order to produce an unsatisfiable Boolean formula, all sequences of truth assignments to the input variables that satisfy the environment assumptions must be considered. This requires one depth-$d$ Boolean unrolling for each possible length-$d$ sequence of inputs, where each unrolling encodes a distinct sequence of inputs in the unrolled Boolean formula. In the worst case, the number of depth-$d$ Boolean formulas that must be generated before an unsatisfiable formula is found grows exponentially in $d$. 

\begin{figure}
\centering
\includegraphics[width=\linewidth]{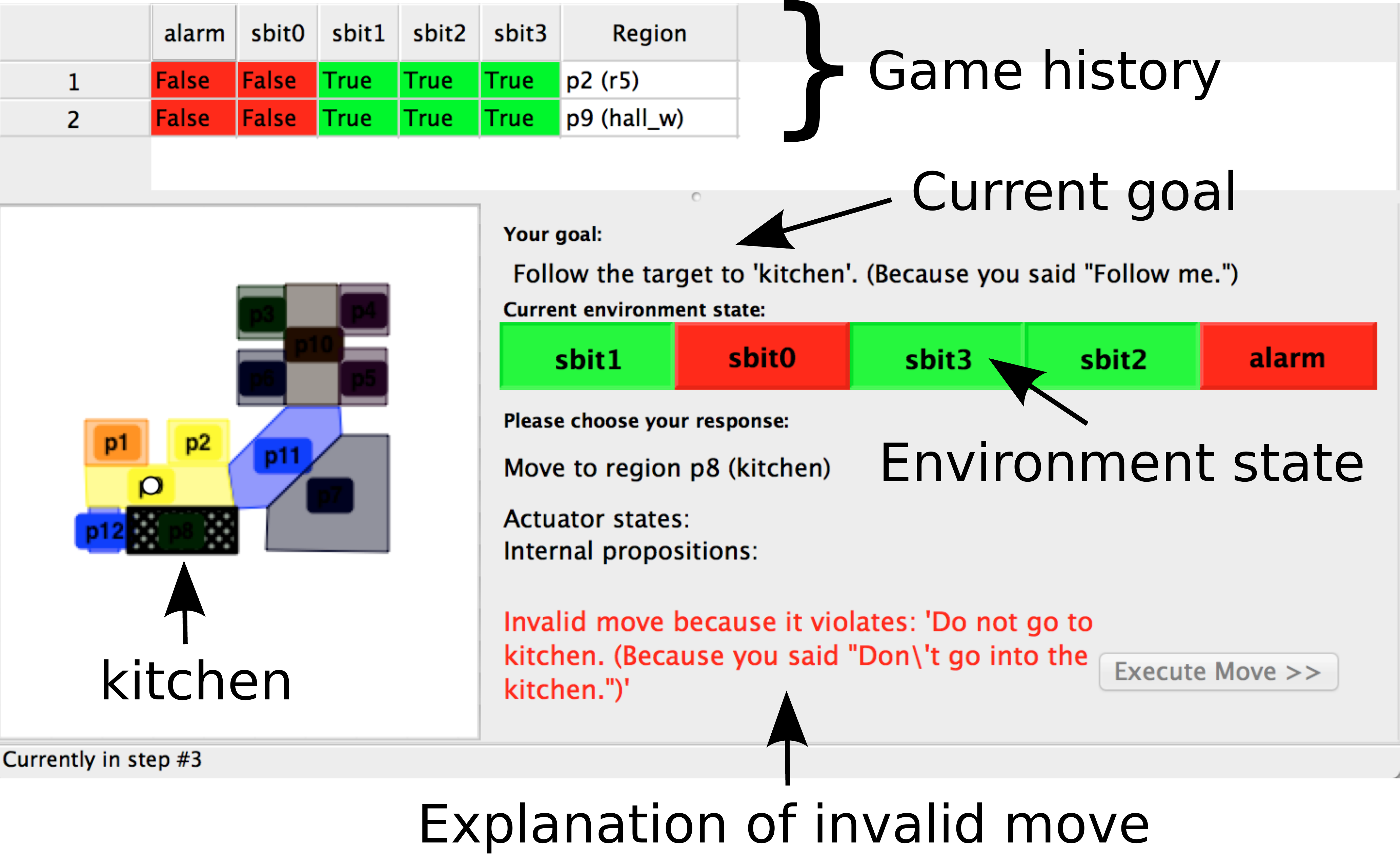}
\caption{Screenshot of interactive visualization tool for Specification~\ref{spec:unreal}.  The user is prevented from following the target into the kitchen in the next step (denoted by the blacked out region) due to the portion of the specification displayed.}
\label{fig:follow_feedback}
\end{figure}

However, unsatisfiable cores do enable a useful enhancement to an interactive visualization of the environment counterstrategy. Since succinctly summarizing the cause of an unrealizable specification is often challenging even for humans, one approach to communicating this cause in a user-friendly manner is through an interactive game (shown in Fig.~\ref{fig:follow_feedback}). The tool illustrates environment behaviors that will cause the robot to fail, by letting the user play as the robot against an adversarial environment. At each discrete time step, the user is presented with the current goal to pursue and the current state of the environment. They are then able to respond by changing the location of the robot and the status of its actuators. Examples of this tool in action are given in \cite{ICRA12,TRO12}.

An initial version of this tool simply prevented the user from making moves that were disallowed by the specification. However, by using the above core-finding technique, a specific explanation can be given about the part of the original specification that would be violated by the attempted invalid move. This is achieved by finding the unsatisfiable core of a single-step satisfiability problem constructed over the user's current state, the desired next state, and all of the robot's specified safety conditions.

Consider Specification~\ref{spec:unreal}, which first appeared in \cite{RSS13}. The robot is instructed to follow a human partner (Line~\ref{spec:unreal:follow}) through the workspace depicted in \ref{fig:hospital_map}. This means that the robot should always eventually be in any room that the human visits. Additionally, the robot has been banned from entering the kitchen in Line~\ref{spec:unreal:avoid}. We discover that the robot cannot achieve its goal of following the human if the human enters the kitchen. This conflict is presented to the user as depicted in Fig. \ref{fig:follow_feedback}:  the environment sets its state to represent the target's being in the kitchen, and then, when the user attempts to enter the kitchen, the tool explains that this move is in conflict with Line~\ref{spec:unreal:avoid}. 

\begin{spec}
\small
\begin{enumerate}
\item Follow me. \label{spec:unreal:follow}
\item Avoid the kitchen. \label{spec:unreal:avoid}
\end{enumerate} 
\caption{Example of unrealizability}
\label{spec:unreal}
\end{spec}

%% file: cores_unreal.tex
As described in Section \ref{mopsy}, the extension of the SAT-based core-finding techniques described in Section \ref{cores_unsat} to unrealizable specifications requires examining the exact environment input sequences that cause the failure. Considering all possible environment input sequences is not feasible; fortunately, the environment counterstrategy sometimes provides us with exactly those input sequences that cause unsynthesizability.  

Consider a counterstrategy $A^e_\varphi = (Q, Q_0, {\cal X}, {\cal Y}, \delta_e, \delta_s, \gamma_{\cal X}, \gamma_{\cal Y}, \gamma_{goals})$ for formula $\varphi$. It allows the following characterizations of deadlock and livelock:

\begin{itemize}
\item
\textbf{Deadlock}
There exists a state in the counterstrategy such that there is a truth assignments to inputs, for which no truth assignment to outputs satisfies the robot transition relation. Formally,
$$\exists q \in Q \mbox{ s.t. } \delta_s(q, \delta_e(q)) = \emptyset$$

\item
\textbf{Livelock}
There exist a set of states $\cal C$ in the counterstrategy such that the robot is trapped in $\cal C$ no matter what it does, and there is some robot liveness $B_k$ in $\varphi_s^g$ that is not satisfied by any state in $\cal C$. Formally,
\[
\begin{array}{l}
\exists {\cal C} \subseteq Q , B_k \preceq \varphi_s^g \mbox{ s.t. } \forall q \in {\cal C}, q \not \models B_k, \delta_s(q, \delta_e(q)) \subseteq {\cal C}
\end{array}
\]

Here ``$q \not \models B_k$'' indicates that the truth assignment given by $\gamma_{\cal X}(q)\cup\gamma_{\cal Y}(q)$ does not satisfy the propositional formula $B_k$. Note that there is always such a set of states in the counterstrategy in the case of livelock.
\end{itemize}

\subsection{Unrealizable Cores for Deadlock}
Consider Specification~\ref{spec:unreal2} on the workspace in Fig. \ref{fig:hallway}. The robot starts in $r5$ with the camera on (\ref{item:b1}). The safety conditions specify that the robot should not pass through the region marked $r5$ if it senses a person (\ref{item:b2}). In addition, the robot must stay in place if it senses a person (\ref{item:b3}). Finally, the robot should always activate its camera (\ref{item:b4}). Here, the environment can force the robot into deadlock by activating the ``person'' sensor ($\pi_{person}$) when the robot is in $r5$, because there is then no way the robot can fulfil both (\ref{item:b2}) and (\ref{item:b3}).

\begin{spec}
\small
\begin{enumerate}
\item Robot starts in r5 with camera ($\varphi_s^i$):\\
\label{item:b1}
$\pi_{r5} \land \pi_{camera}$

\item If you are sensing person then do not r5 ($\varphi_s^t$):\\
\label{item:b2}
$\G (\X \pi_{person} \Rightarrow \neg \X \pi_{r5})$

\item If you are sensing person then stay there ($\varphi_s^t$):\\
\label{item:b3}
$\G (\X \pi_{person} \Rightarrow (\X \pi_{start} \Leftrightarrow \pi_{start} \land \X \pi_{r2} \Leftrightarrow \pi_{r2}...))$

\item Always activate your camera ($\varphi_s^t$):\\
\label{item:b4}
$\G \X \pi_{camera}$
\end{enumerate}
\caption{Core-finding example -- deadlock}
\label{spec:unreal2}
\end{spec}

The environment counterstrategy $A^e_\varphi$ is as follows:

\begin{itemize}
\item $Q = Q_0 = \{q_1\}$
\item ${\cal X} = \{\pi_{person}\}$, ${\cal Y} = \{\pi_{r5}, \pi_{camera}\}$ 
\item $\delta_e(q_1) = \{\pi_{person}\}$, $\delta_s(q_1, \{\pi_{person}\}) = \emptyset$, $\delta_s(q_1, \emptyset) = \emptyset$
\item $\gamma_{\cal X}(q_1) = \{\pi_{person}\},~\gamma_{\cal Y}(q_1) = \{\pi_{r5},\pi_{camera}\}$
\item $\gamma_{goals}(q_1) = 1$
\end{itemize}

The state $q_1$ is deadlocked, because given the input $\pi_{person}$ in the next time step, there is a conflict between safety conditions \ref{item:b2} and \ref{item:b3}, and the robot has no valid move (so $\delta_s(q_1, \{\pi_{person}\}) = \emptyset$). Note that $\delta_s(q_1, \emptyset) = \emptyset$ indicates that the environment strategy does not include any transition out of $q_1$ where the environment does not activate the ``person'' sensor.

For $q \in Q$, the \emph{propositional-representation of q} is defined as:
$$\psi_{state}(q) = \bigwedge_{x \in \gamma_{\cal X}(q)} x \land \bigwedge_{x \in {\cal X}\backslash\gamma_{\cal X}(q)} \neg x \land \bigwedge_{y \in \gamma_{\cal Y}(q)} y \land \bigwedge_{y \in {\cal Y}\backslash\gamma_{\cal Y}(q)} \neg y$$

In the example above, $\psi_{state}(q_1) = \pi_{person} \land \pi_{r5} \land \pi_{camera}$.

As before, let $\pi^i$ represent the value of $\pi \in AP$ at time step $i$, and $AP^i = \{\pi^i \mid \pi \in AP\}$. For example, in Specification~\ref{spec:unreal2}, $AP^0 = \{\pi_{person}^0,\pi_{r5}^0, \pi_{camera}^0\}$ and $AP^1 = \{\pi_{person}^1,\pi_{r5}^1, \pi_{camera}^1\}$. 

Given LTL specification $\varphi$,  $q \in Q$ such that $\delta_s(q, \delta_e(q)) = \emptyset$, construct a propositional formula over $AP^0 \cup AP^1$ as follows:
\[
\begin{array}{l}
\psi_{dead}(q, \varphi) =
\psi_{state}(q)[\pi/\pi^0]
\displaystyle\land \bigwedge_{z \in \delta_e(q)} z^1 \land \bigwedge_{z \in {\cal X}\backslash\delta_e(q)} \neg z^1 \\
~~~~~~~~~~~~~~~\land \varphi_s^t[\X \pi/\pi^1][\pi/\pi^0],
\end{array}
\]

Intuitively, this formula represents the satisfaction of the robot safety condition in the next step from state $q$, with the additional restriction that the input variables be bound to the values provided by $\delta_e(q)$ in the next time step. 
 
In the above case, $\psi_{dead}(q_1, \varphi) =$
\[
\begin{array}{l}
\pi_{person}^0 \land \pi_{r5}^0 \land \pi_{camera}^0 \land \pi_{person}^1 \land \pi_{camera}^1
\land \varphi_{topo}^0\\
\land (\pi_{person}^1 \Rightarrow \neg \pi_{r5}^1)
\land (\pi_{person}^1 \Rightarrow (\pi_{r5}^1 \Leftrightarrow \pi_{r5}^0 \land ...)),
\end{array}
\]
where $\varphi_{topo}^i$ is a formula over $AP^i \cup AP^{i+1}$ representing the topological constraints on the robot motion at time $i$ (i.e. which rooms it can move to at time $i+1$ given where it is at time $i$, and mutual exclusion between rooms).

Note that if $q$ is a deadlocked state, then by definition $\psi_{dead}(q, \varphi)$ is unsatisfiable, since there is no valid setting to the robot propositions in the next time step starting from q. A SAT solver can now be used to find a minimal unsatisfiable subformula, as in Section \ref{cores_unsat}.

In the above example, the SAT solver finds the core of $\psi_{dead}(q, \varphi)$ as the subformula
\[
\begin{array}{l}
\pi_{r5}^0 \land \pi_{person}^1
\land (\pi_{person}^1 \Rightarrow \neg \pi_{r5}^1) 
\land (\pi_{person}^1 \Rightarrow (\pi_{r5}^1 \Leftrightarrow \pi_{r5}^0)).
\end{array}
\]

This is because the two statements combined require the robot to both stay in r5 and not be in r5 in time step 1. This gives us a core explanation of the deadlock caused in state $q$. Taking the union over the cores for all the deadlocked states provides a concise explanation of how the environment can force the robot into a (not necessarily unique) deadlock situation. Section \ref{examples} contains another, more complex example demonstrating unrealizable core-finding for deadlocked specifications.

\subsection{Unrealizable Cores for Livelock}
Consider Specification~\ref{spec:unreal1} again. If the environment action is to always set $\pi_{person}$, then the safety requirement in \ref{item:a1} enforces that the robot will never activate $\pi_{r5}$, because it is explicitly forbidden from doing so when sensing a person. This is livelock because the robot can continue to move between $start$ and $r2-r4$. The environment counterstrategy $A^e_\varphi$ is as follows:

\begin{itemize}
\item $Q = \{q_1, q_2, q_3, q_4\},~Q_0 = \{q_1\}$
\item ${\cal X} = \{\pi_{person}\},{\cal Y} = \{\pi_{start}, \pi_{r2}, \pi_{r3}, ..., \pi_{r8}, \pi_{goal}, \pi_{camera}\}$ 
\item $\forall q \in Q, \delta_e(q) = \{\pi_{person}\}$
\item 
\[\delta_s(q_i, \{\pi_{person}\}) =
\left\{
\begin{array}{ll}
	\{q_1, q_2\} & \mbox{ if } i = 1\\
	\{q_{i-1}, q_i, q_{i+1}\}&\mbox{ for } 2 \le i \le 3\\
	\{q_3, q_4\}&\mbox{ if } i = 4
\end{array}\right.\]

Additionally,  $\delta_s(q, \emptyset) = \emptyset$ for all $q$.
\item $\forall q \in Q, \gamma_{\cal X}(q) = \{\pi_{person}\}$. 
\item
\[\gamma_{\cal Y}(q_i) =
\left\{
\begin{array}{ll}
	\{\pi_{camera}, \pi_{start}\} & \mbox{ if } i = 1\\
	\{\pi_{camera}, \pi_{r_i}\}&\mbox{ for } 2 \le i \le 4
\end{array}\right.\]
\item $\forall i, \gamma_{goals}(q_i) = 1$ (since there is only one goal).
\end{itemize}

\subsubsection{Countertraces}\label{sec:countertraces}
One of the main sources of complexity in analyzing an environment counterstrategy is that it might depend on the behavior of the system. However, sometimes it is possible to extract a single trace of inputs such that no output trace fulfil the specification. Following \cite{Konig09}, we call such a trace a \emph{countertrace}. A countertrace does not always exist, but it does simplify our analysis. Computing countertraces is expensive \cite{Konig09}, but it is possible to identify whether a given counterstrategy is a countertrace by checking that all paths from an initial state follow the same sequence of environment inputs. For the analysis that follows, we assume that the counterstrategy is a countertrace. We will later discuss how to analyze counterstrategies that are not traces. Note that counterstrategy $A^e_\varphi$ for Specification \ref{spec:unreal1} above is a countertrace. 

In the case of livelock, we know there exists a set of states $\cal C$ in the counterstrategy that trap the robot, locking it away from the goal. Without loss of generality, $\cal C$ consists of (possibly overlapping) cycles of states.  In the specifications of the form considered in this work, robot goals are of the form $\G \F B_i$ for $1 \le i \le n$, where each $B_i$ is a propositional formula over $AP = {\cal X} \cup {\cal Y}$. Suppose the algorithm in \cite{TRO12} identified goal $\G \F B_k$ as the goal responsible for livelock. Let $Q_k$ be the set of all states in $A^e_\varphi$ that prevent goal $B_k$, and let ${\cal C}_k$ be the set of \emph{maximal $k$-preventing cycles} in $Q_k$, i.e. cycles that are not contained in any other cycle in $Q_k$ (modulo state-repetition). Let $C_1 = (q^1_0, q^1_1,...,q^1_a)$ and $C_2 = (q^2_0, q^2_1,...,q^2_b)$ be cycles, and define $C_1 \prec C_2$ if $a < b$ and there is some offset index $o$ in $C_2$ such that all of $C_1$ is found in $C_2$ starting at $o$, i.e. $q^1_i = q^2_{(i + o)\bmod{(b+1)}}$ for all $0 \le i \le a$. This expresses that $C_1$ is a strict sub-cycle of $C_2$. Formally,
\[
\begin{array}{rll}
Q_k &=& \{q \in Q | \gamma_{goals}(q) = k\}\\\\
{\cal C}^{all}_k &=& \{(q_0,q_1,...,q_l) | \forall 0 \le i \le l, q_i \in Q_k,\\
~~~~~~~~~~~~~~~~&&
\begin{array}{l}
	~\forall i <l,q_{i+1} \in \delta_s(q_i, \delta_e(q_i)), q_i \ne q_{i+1},\\
	~~q_{0} \in \delta_s(q_l, \delta_e(q_l)), q_0 \ne q_l\},
\end{array}\\\\
{\cal C}_k &=& \{C \in C^{all}_k | \forall C' \in C^{all}_k, C \not \prec C'\}.
\end{array}
\]

In Specification \ref{spec:unreal1}, there is only one goal, $\G \F \pi_{goal}$. $C_1 = (q_1,q_2,q_3,q_4,q_3,q_2)$ is a maximal $1$-preventing cycle. 

Given an initial state $q$, a depth $d$ and an LTL safety formula $\varphi_s^t$ over $\pi \in AP$, we construct a propositional formula $\psi^d(\varphi_s^t,q)$ over $\bigcup_{0 \le i \le d+1}AP^i$ as:
$$\psi^d(\varphi_s^t,q) = \psi_{state}(q)[\pi/\pi^0] \land \bigwedge_{0 \le i \le d} \varphi_s^t[\X \pi/\pi^{i+1}][\pi/\pi^i].$$ This formula is called the \emph{depth-$d$ unrolling of $\varphi_s^t$ from $q$}, and represents the tree of length-$d+1$ truth assignment sequences that satisfy $\varphi_s^t$, starting from $q$. Note that a depth-$d$ unrolling governs $d+1$ time steps, because each conjunct in $\varphi_s^t$ governs the current as well as next time steps. In the example, $\psi^d(\varphi_s^t,q_1) =$ 
\[
\begin{array}{l}
\tiny
\displaystyle\pi_{start}^0 \land \pi_{camera}^0 \land \bigwedge_{0 \le i \le d} (\varphi_{topo}^i \land \pi_{camera}^{i+1}
\land (\pi_{person}^{i+1} \Rightarrow \neg \pi_{r5}^{i+1})).
\end{array}
\]

Given a cycle of states $C = (q_0, q_1, ..., q_l) \in A^e_\varphi$, and a depth $d$, construct a propositional formula $\psi^d_{\cal X}(C)$ over $\bigcup_{0 \le i \le d}{\cal X}^i$, where  $x^i \in {\cal X}^i$ represents the value of each input $x \in {\cal X}$ in state $q_{i\bmod{(l+1)}}$ for $0 \le i \le d$, as: 
$$\psi^d_{\cal X}(C) = \bigwedge_{0 \le i \le d} (\bigwedge_{p \in \gamma_{\cal X}(q_{i \bmod{(l+1)}})}x^i \land \bigwedge_{p \in {\cal X}\backslash\gamma_{\cal X}(q_{i \bmod{(l+1)}})}\neg x^i).$$
This formula is called the \emph{depth-$d$ environment-unrolling of $C$}, and represents the sequence of inputs seen when following cycle $C$ for $d$ time-steps. In the example, the depth-$d$ environment unrolling of $C_1$ is $\psi^d_{\cal X}(C) = \bigwedge_{0 \le i \le d} \pi_{person}^i$. 

Now, given an LTL safety specification $\varphi_s^t$ over $\pi \in AP$, a goal $B_k$, a maximal $k$-preventing cycle $C_k = (q_0,q_1,...,q_l) \in {\cal C}_k$, and an unrolling depth $d$, construct propositional formula $\psi^d_{live}(B_k, C_k, \varphi_s^t)$ over $\bigcup_{0 \le i \le d+1}AP^i$ as:
\[
\begin{array}{l}
\psi_{live}^d(B_k, C_k, \varphi_s^t) =\\
~~~~~~~~~~~~~~~\psi^{d+1}_{\cal X}(C_k) \land \psi^d(\varphi_s^t, q_0) \land B_k[\pi/\pi^{d}].$$
\end{array}
\]

Intuitively, this formula expresses the requirement that the goal $B_k$ be fulfilled after some depth-$d$ unrolling of the safety formula starting from state $q_0$, given the input sequence provided by $\psi^{d+1}_{\cal X}(C_k)$ (note that this input sequence extends to the final time step in the safety formula unrolling). This is an unsatisfiable propositional formula, and can be used to determine the core set of clauses that prevent a goal from being fulfilled. Taking the union of cores over all $C_k \in {\cal C}_k$ gives a concise explanation of the ways in which the environment can prevent the robot from fulfilling the goal. This step makes use of the fact that the counterstrategy is a countertrace, since otherwise the reason for unrealizability might involve an interplay between two input sequences. 

In the above example, $\psi^d_{live}(\pi_{goal}, C_1, \varphi_s^t) =$
\[
\begin{array}{l}
\tiny
\displaystyle\pi_{start}^0 \land \pi_{camera}^0 \land \bigwedge_{0 \le i \le d+1} \pi_{person}^{i}  \\
~~\displaystyle\land \bigwedge_{0 \le i \le d} (\varphi_{topo}^i \land \pi_{camera}^{i+1} \land (\pi_{person}^{i+1} \Rightarrow \neg \pi_{r5}^{i+1})) \\
~~~\displaystyle\land \pi_{goal}^{d}.
\end{array}
\]

In the case of livelock, the choice of unroll depth $d$ determines the quality of the core returned. Recall that for deadlock, the propositional formula $\psi_{dead}(q, \varphi)$ is built over just one step, since it is already known to cause a conflict with the robot transition relation, and be unsatisfiable. The unsatisfiable core of this formula is a meaningful unrealizable core in this case because it provides the immediate reason for the deadlock. For livelock, on the other hand, determining the shortest depth to which a cycle $C_k$ must be unrolled to produce a meaningful core is not obvious. 

In the above example, for unroll depths less than or equal to 8, the unsatisfiable core returned will include just the environment topology, since the robot cannot reach the goal from the start in 8 steps or fewer, even if it is allowed into $r5$; however, this is not a meaningful core. Unrolling to depth 9 or greater returns the expected subformula that includes $\bigwedge_{0 \le i \le d} (\pi_{person}^{i+1} \Rightarrow \neg \pi_{r5}^{i+1})$. Automatically determining the shortest depth that will produce a meaningful core remains a research challenge, but a good heuristic is to use the maximum distance between two states in the environment counterstrategy (i.e. the diameter of the graph representing the counterstrategy, or the sum of the diameters of its connected components). 

\subsubsection{General Counterstrategies}
It may be tempting to try and use a similar approach to find unrealizable cores from counterstrategies that are not countertraces. However, since the input sequences can now depend on the system behavior, it becomes necessary to encode all possible paths through the counterstrategy. Even so, since the counterstrategy only contains paths that are valid for the system specification, we found that the returned core often contained these added constraints on the system instead of the original specification. Instead, we used the approach described in Section \ref{cores_iterate} to extract a minimal core in the case of livelock, when the counterstrategy was not a countertrace.

\begin{algorithm}
\caption{Unrealizable Cores via SAT solving}  \label{alg:unreal_bmc}
\begin{algorithmic}[1]
\small
\makeatletter\setcounter{ALG@line}{0}\makeatother
\Function{\tt UNREAL\_BMC}{$\varphi$, reason}
	\State  MUS $\gets \emptyset$
	\State  $A^e_\varphi = (Q, Q_0, ...,) \gets$ {\tt COUNTERSTRATEGY}($\varphi$)
	\If {reason==deadlock}
	\For {$q \in Q$ such that $\delta_s(q, \delta_e(q)) = \emptyset$}
      	\State MUS $\gets$ MUS $\cup$ {\tt SAT\_SOLVER}($\psi_{dead}(q, \varphi$))
	\EndFor
	\Else 
	\State $k \gets$ livelocked goal
	\If{{\tt IS\_COUNTERTRACE}($A^e_\varphi$, $k$)}
	\State $C^{all}_k$ $\gets$ {\tt FIND\_PREVENTING\_CYCLES}($A^e_\varphi$, $k$)
	\For {${\cal C}_k \in \{C \in C^{all}_k | \forall C' \in C^{all}_k, C \not \prec C'\}$}
	\State MUS $\gets$ MUS $\cup$ {\tt SAT\_SOLVER}($\psi^d_{live}(B_k, C_k, \varphi_s^t)$) 
	\EndFor
	\Else
	\State $\varphi_s^{badInit} \gets$ {\tt CHOOSE\_ONE}($Q_0$)
	\State MUS $\gets$ {\tt UNREAL\_ITERATE}({$\varphi, \varphi_s^{badInit}, k$})
	\EndIf
	\EndIf
    \Return {\tt MAP\_BACK}(MUS)
 \EndFunction
\end{algorithmic}  
\end{algorithm}

Algorithm \ref{alg:unreal_bmc} summarizes the core-finding procedure described in this section. The module {\tt IS\_COUNTERTRACE} checks that all paths form a single initial state in the counterstrategy follow the same sequence of inputs. {\tt FIND\_PREVENTING\_CYCLES} finds cycles in $A^e_\varphi$ that prevent goal $k$. {\tt UNREAL\_ITERATE} is described in Section \ref{cores_iterate}.

Note that, since unsatisfiability is a special case of unrealizability (in which not just \emph{some}, but \emph{any} environment can prevent the robot from fulfilling its specification), the above analysis also applies to unsatisfiable specifications. Moreover, a countertrace can always be extracted for an unsatisfiable specification, and so the approach in Section \ref{sec:countertraces} applies for livelock. However, the analysis presented in Section \ref{cores_unsat} is more efficient for unsatisfiability, as it does not require explicit-state extraction of the environment counterstrategy.

%% file: cores_iterate.tex
As discussed in Sections \ref{cores_unsat} and \ref{cores_unreal}, the SAT-based approach to identifying an unsynthesizable core for the case of livelock presents the challenge of determining a depth to which the LTL formula must be instantiated with propositions. This minimal depth is often tied to the number of regions in the robot workspace, and is usually easy to estimate. However, no efficient, sound method is known for determining this minimal unrolling depth. In addition, if the counterstrategy is not a countertrace, the techniques in Section \ref{cores_unreal} do not readily apply to extracting an unrealizable core. In both these cases, the SAT-based analysis described in Section \ref{cores_unreal} may return a core that does not capture the real cause of failure, causing confusion when presented to the user. Fortunately, alternative, more computationally expensive techniques can be used to return a minimal core in these cases.

This section presents one such alternative approach to determining the minimal subset of the robot safety conjuncts that conflicts with a specified goal. The approach is based on iterated realizability checks, removing conjuncts from the safety formula and testing realizability of the remaining specification. While this approach is guaranteed to yield a minimal unsynthesizable core, it requires repeated calls to a realizability oracle, which may be expensive for specifications with a large number of conjuncts.

Recall from Section \ref{prelims} the syntactic form of the LTL specifications considered in this work. In particular, the formula $\varphi_s^g$ is a conjunction $\bigwedge_{j=1}^{n^g_s} \G \F B_j$, where each $B_j$ is a Boolean formula over $AP$, and represents an event that should occur infinitely often when the robot controller is executed. Similarly, $\varphi_s^t$ represents the robot safety constraints; it is a conjunction $\bigwedge_{i=1}^{n^t_s} \G A_i$ where each $A_i$ is a Boolean formula over $AP$ and $\X AP$.

In the case of livelock, the initial specification analysis presented in \cite{TRO12} provides a specific liveness condition $B_k$ that causes the unsynthesizability (i.e. either unsatisfiability or unrealizability), and can also identify one of the initial states $\varphi_s^{badInit}$ from which the robot cannot fulfil $B_k$. However, the specific conjuncts of the safety formula $\varphi_s^t$ that prevent this liveness are not identified. The key idea behind using realizability tests to determine an unrealizable or unsatisfiable core of safety formulas is as follows. If on removing a safety conjunct from the robot formula, the specification remains unsynthesizable, then there exists an unsynthesizable core that does not include that conjunct (since the remaining conjuncts are sufficient for unsynthesizability). Therefore, in order to identify an unsynthesizable core, it is sufficient to iterate through the conjuncts of $\varphi_s^t$, removing safety conditions one at a time and checking for realizability. 

Algorithm \ref{alg:iterate} presents the formal procedure for performing these iterated tests, given the index $k$ of the liveness condition that causes the unsynthesizability. Denote by $\varphi_s[S,\varphi_s^{badInit},k] \subseteq \varphi_s$ the formula $\varphi_s^{badInit} \land \bigwedge_{i \in S} \G A_i \land \G\F B_k$ for indices in a set $S$.  Let $S_i$ denote set $S$ at iteration $i$. Set $S_1$ is initialized to the indices of all safety conjuncts, i.e. $S_1 = \{1,...,n_s^t\}$ in line 2. In each iteration of the loop in lines 3-7, the next conjunct $A_i$ is omitted from the robot transition relation, and realizability of $\varphi_e \Rightarrow \varphi_s[S_i\backslash\{i\},\varphi_s^{badInit},k]$ is checked (line 4). If removing conjunct $i$ causes an otherwise unsynthesizable specification to become synthesizable, it is retained for the next iteration (line 5); otherwise it is permanently deleted from the set of conjuncts $S_i$ (line 6-7). After iterating through all the conjuncts in $\{1,...,n_s^t\}$, the final set $S_{n_s^t+1}$ determines a minimal unsynthesizable core of $\varphi_e \Rightarrow \varphi_s$ that prevents liveness $k$. Note that the core is non-unique, and depends both on the order of iteration on the safety conjuncts, and on the initial state $\varphi_s^{badInit}$ returned by the synthesis algorithm.

\begin{theorem}
Algorithm \ref{alg:iterate} yields a minimal unsynthesizable core of $\varphi_e \Rightarrow \varphi_s$. \end{theorem}

\textit{Proof:}
Each iteration $i$ of the loop in Algorithm \ref{alg:iterate}, lines 3-7, maintains the invariant that $\varphi_e \Rightarrow \varphi_s[S_i,\varphi_s^{badInit},k]$ is unsynthesizable; thus, $\varphi_e \Rightarrow \varphi_s[S_{n_s^t+1},\varphi_s^{badInit},k]$ is unsynthesizable when the loop is exited. 

Moreover, removing any of the safety conjuncts in $S_{n_s^t+1}$ yields a synthesizable specification. To see this, assume for a contradiction that there exists $j \in S_{n_s^t+1}$ such that $\varphi_e \Rightarrow \varphi_s[S_{n_s^t+1}\backslash\{j\},\varphi_s^{badInit},k]$ is unsynthesizable. Clearly, $S_{n_s^t+1} \subseteq S_j$, so by definition of $\preceq$, $\varphi_s[S_{n_s^t+1}\backslash\{j\},\varphi_s^{badInit},k] \preceq \varphi_s[S_{j}\backslash\{j\},\varphi_s^{badInit},k]$. Therefore, if $\varphi_e \Rightarrow \varphi_s[S_{j}\backslash\{j\},\varphi_s^{badInit},k]$  is synthesizable, then $\varphi_e \Rightarrow \varphi_s[S_{n_s^t+1}\backslash\{j\},\varphi_s^{badInit},k]$  must be synthesizable, since any implementation that satisfies $\varphi_s[S_{j}\backslash\{j\},\varphi_s^{badInit},k]$  also satisfies $\varphi_s[S_{n_s^t+1}\backslash\{j\},\varphi_s^{badInit},k]$ . Since $j$ was not removed from $S_j$ on the $j^{th}$ iteration, $\varphi_e \Rightarrow \varphi_s[S_{j}\backslash\{j\},\varphi_s^{badInit},k]$ is synthesizable. It follows that $\varphi_e \Rightarrow \varphi_s[S_{n_s^t+1}\backslash\{j\},\varphi_s^{badInit},k]$ must be synthesizable, a contradiction.

\begin{algorithm}
\caption{Unsynthesizable Cores via Iterated Realizability Testing} \label{alg:iterate}
\begin{algorithmic}[1]
\small
\makeatletter\setcounter{ALG@line}{0}\makeatother
	\Function{UNREAL\_ITERATE}{$\varphi = \varphi_e, \Rightarrow \varphi_s, \varphi_s^{badInit}, k$}
	\State $S_1 = \{1,2,...,n_s^t\}$
    \For {$i := 1$ to $n^t_s$ }
      \If {$(\varphi_e \Rightarrow \varphi_s[S_i\backslash\{i\},\varphi_s^{badInit},k])$ is synthesizable}
      	\State $S_{i+1} \leftarrow S_i$
      \Else
      	\State $S_{i+1} \leftarrow S_i\backslash\{i\}$
      \EndIf
    \EndFor
    \Return $\varphi_s[S_{n_s^t+1},\varphi_s^{badInit},k])$
    \EndFunction
\end{algorithmic}  
\end{algorithm}

Note that Algorithm \ref{alg:iterate} yields an unsynthesizable core for livelock, for both unsatisfiable and unrealizable specifications. It is sound and complete, because it will always yield a minimal set of safety conditions that prevent the relevant liveness.  As compared with the methods presented in Sections \ref{cores_unsat} and \ref{cores_unreal}, it circumvents the problem of determining the depth to which to instantiate the LTL safety formula in a propositional SAT instance. Moreover, if $\varphi_s[S,\varphi_s^{badInit},k]$ is replaced with $\varphi_s^{badInit} \land \bigwedge_{i \in S} \G A_i \land \G\F \mbox{TRUE}$ (i.e. the robot liveness condition is trivial), the algorithm also yields an unsynthesizable core in the case of deadlock.  

\subsection{Computational Tradeoffs}
There is a computational tradeoff involved in performing a synthesizability (i.e. realizability) check once for every conjunct in the  safety formula, instead of once for the entire specification. Algorithm \ref{alg:iterate} checks synthesizability once in each iteration of the loop in lines 3-7. Using the efficient algorithm in \cite{Piterman06}, each realizability check takes time $O((mn\Sigma)^3)$, where $\Sigma$ is the size of the state space, i.e. $\Sigma = 2^{|\cal X \cup Y|}$, and $m,n$ are the number of environment and system liveness conditions, respectively. Therefore the complexity of Algorithm \ref{alg:iterate} is $O((n_s^t)(mn\Sigma)^3)$. On the other hand, the complexity of the approach described in Section \ref{cores_unreal} requires only one call to the counterstrategy synthesis algorithm, but multiple calls to the SAT solver. The SAT solver is invoked with Boolean formulas in CNF form that are, in the worst case, exponential in the size of the original LTL conjuncts, although the conjunctive form of the original GR(1) specifications mitigates some of this blowup in practice. Additionally, iterated realizability tests do not require explicit extraction of the environment counterstrategy, as in the case of the SAT-based tests presented in Section \ref{cores_unreal}. The relative appropriateness of the two methods (SAT-based vs. iterated realizability testing) for the case of deadlock will depend on the specific unsynthesizable formula.

For example, in Specification \ref{spec:unreal}, the calls to {\tt SAT\_SOLVER} made as part of the approach in Algorithm \ref{alg:unreal_bmc} took $<$1ms each to complete on a 1.3 GHz Intel Core i5 processor with 8GB of RAM. On the other hand, for the same specification with line (\ref{item:a1}) replaced with \textit{Always do not r5} ($\G (\neg \X \pi_{r5}))$, each call took approximately 60s. This was exceptional, and most of the calls to {\tt SAT\_SOLVER} for the examples in this paper (including those in section \ref{examples}) took $<$1ms each. On the other hand, each call to {\tt COUNTERSTRATEGY} for the approach in Algorithm \ref{alg:iterate} took approximately 5ms for the example in Section \ref{example:livelock}.

%% file: examples.tex
This section presents examples of the cores identified for unsynthesizable specifications. The examples presented previously appeared in \cite{ICRA12}, and this section demonstrates the improvement of the proposed approach over the analysis presented in that work.  These examples were run using the open-source LTLMoP toolbox \cite{LTLMoP}, within which all the technical outcomes presented in this paper have been implemented.

\begin{figure}
\centering
\includegraphics[width=.25\textwidth]{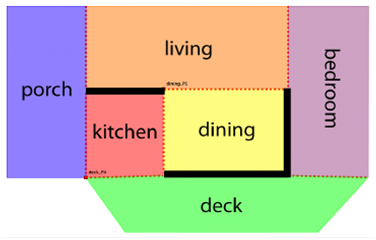}
\caption{Map of robot workspace for specifications in Section \ref{examples}}
\label{fig:firefighting}
\end{figure}

\begin{figure*}[htp!]
\captionsetup{width=0.5\textwidth} 
\centering
\subfigure[Sentences highlighted using approach in \cite{ICRA12}]{
\includegraphics[width=0.45\textwidth]{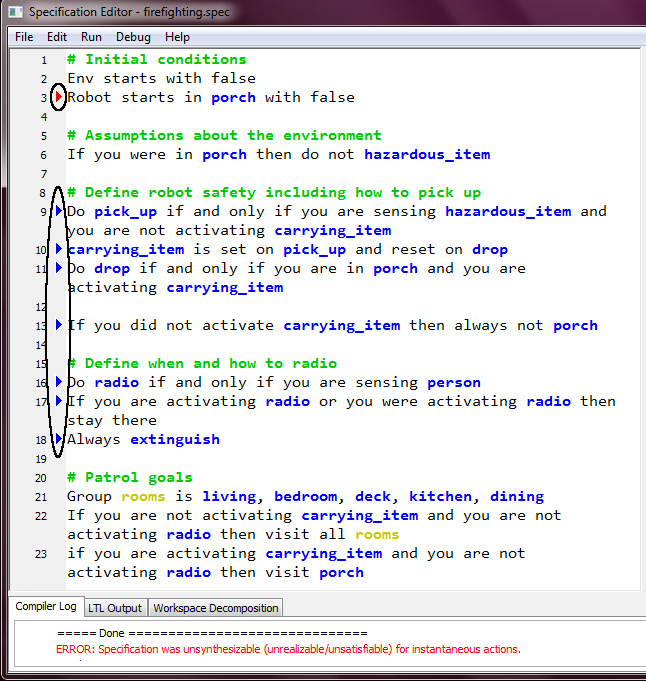}
\label{fig:deadOld}
}
\subfigure[Sentences highlighted using proposed approach]{
\includegraphics[width=0.45\textwidth]{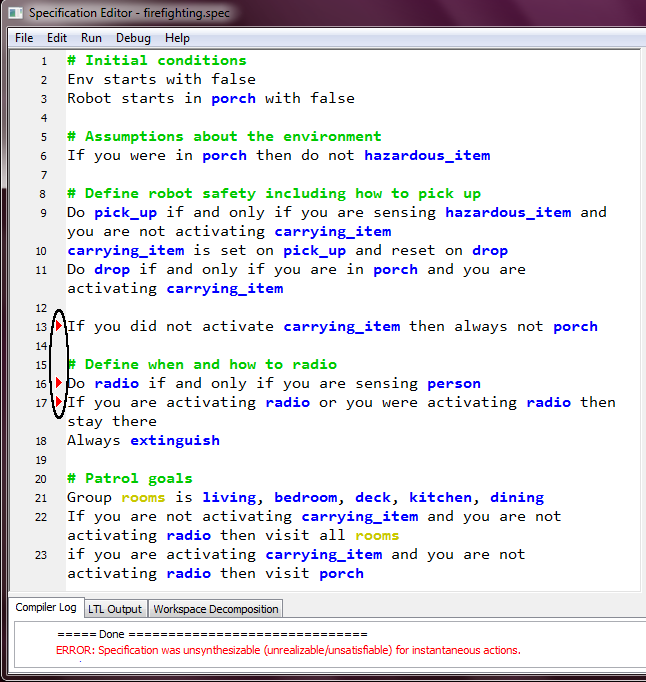}
\label{fig:deadNew}
}
\caption{Core-Finding Example: Deadlock}
\label{fig:deadlock}
\end{figure*}

\begin{figure*}[htp!]
\captionsetup{width=0.5\textwidth} 
\centering
\subfigure[Sentences highlighted using approach in \cite{ICRA12}]{
\includegraphics[width=0.44\textwidth]{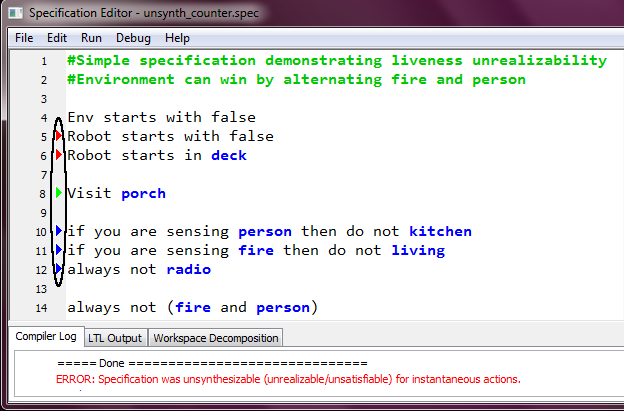}
\label{fig:liveOld}
}
\subfigure[Sentences highlighted using proposed approach]{
\includegraphics[width=0.44\textwidth]{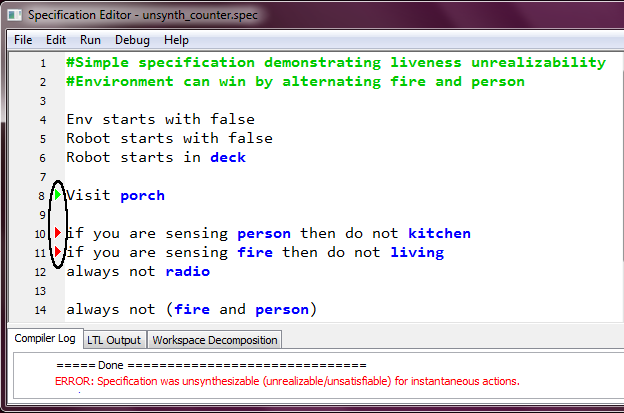}
\label{fig:liveNew}
}
\caption{Core-Finding Example: Livelock}
\label{fig:livelock}
\end{figure*}

\subsection{Deadlock}
Consider the specification in Fig. \ref{fig:deadlock}, where the robot is operating in the workspace depicted in Fig. \ref{fig:firefighting}. The robot starts in the porch. The safety conditions govern what it should do when it senses a ``person'' (stay with them and radio for help) or a ``hazardous item'' (pick up the hazardous item and carry it to the porch). The robot should not visit the porch unless it is carrying a hazardous item. The robot's goals are to patrol all rooms in the workspace. 

The environment can cause deadlock by setting the person sensor to true and the hazardous item sensor to false when the robot is in the porch. Note that sensing a hazardous item results in the robot activating the ``pick\_{up}'' action, which in turn results in the proposition ``carrying\_{item}'' being set. Similarly, sensing a person results in the robot turning on the radio. Now the state in which both ``radio'' and ``carrying\_{item}'' are true in the porch is a deadlocked state because of the safety conditions, ``If you are activating radio or you were activating radio then stay there'' and ``If you did not activate carrying\_{item} then always not porch'', since there is no way to satisfy both from this state.

Fig. \ref{fig:deadOld} depicts the sentences highlighted by the algorithm in \cite{ICRA12}. A subset of sentences in the specification is identified by triangle-shaped markers in the left-hand margin, and the color-coding is based on whether they correspond to initial, safety or liveness conditions. The sentences highlighted in \ref{fig:deadOld} include all initial (red) and safety (blue) conditions, which forms a very large subset of the original specification. On the other hand, Fig. \ref{fig:deadNew} depicts the much smaller subset of guilty sentences returned by the analysis presented in Algorithm \ref{alg:unreal_bmc} (these sentences are all highlighted in red). The core sentences highlighted 
correspond to the safety conditions that cause deadlock -- in this example, removing any one of these sentences results in a synthesizable specification.

\subsection{Livelock}\label{example:livelock}

Consider the specification in Fig. \ref{fig:livelock}, also in the same workspace. The robot starts in the deck and its goal is to visit the porch. However, based on whether it senses a person or a fire, it has to keep out of the kitchen and the living room, respectively. 
%
%
%
%
%
Fig. \ref{fig:liveOld} depicts the sentences highlighted by the algorithm in \cite{ICRA12}, which includes all safety conditions (red) in addition to the goal (green). This includes irrelevant sentences, such as the one that requires the robot to always turn on the camera. Fig. \ref{fig:liveNew} depicts the core returned by Algorithm \ref{alg:iterate} -- only those safeties that directly contribute to keeping the robot out of the porch are returned.

%% file: conclusions.tex
This paper provides techniques for analyzing high-level robot specifications that are unsynthesizable, with the goal of providing a minimal explanation for why the robot specification is inconsistent, or how the environment can prevent the robot from fulfilling the desired guarantees. The causes of failure presented in this work take the form of unsynthesizable subsets of the original specification, or cores. A suite of SAT-based techniques is presented for identifying unsatisfiable and unrealizable cores in the case of deadlock and most cases of livelock; iterated realizability checking is used to identify cores in cases where the SAT-based analysis fails. Examples show that the additional analysis provides improvements in terms of reducing the number of sentences in the original specification highlighted, and ignoring irrelevant subformulas. 

Future work includes automatically determining the depth for obtaining a meaningful core in the case of livelock for the SAT-based approaches, and exploring SAT-based techniques that do not require explicit state extraction of the counterstrategy automaton. Another direction for future study is a systematic empirical comparison of SAT-based techniques with approaches based on iterated realizability testing, to evaluate scalability and relative computation time for practical examples. 